\documentclass[lettersize,journal]{IEEEtran}
\usepackage{amsmath,amsfonts}
\usepackage{algorithmic}
\usepackage{algorithm}
\usepackage{array}
\usepackage[caption=false,font=normalsize,labelfont=sf,textfont=sf]{subfig}
\usepackage{textcomp}
\usepackage{stfloats}
\usepackage{url}
\usepackage{verbatim}
\usepackage{graphicx}
\usepackage{cite}
\usepackage{xcolor}         
\usepackage{makecell}
\usepackage{colortbl}
\usepackage{pifont}
\usepackage[utf8]{inputenc} 
\usepackage[T1]{fontenc}    
\usepackage{multirow}
\usepackage{tabularx}
\usepackage{array}
\usepackage{amsmath}
\usepackage{booktabs} 
\usepackage{svg}
\begin{document}

\title{RainBalance: Alleviating Dual Imbalance in GNSS-based Precipitation Nowcasting via Continuous Probability Modeling}

\author{
Yifang~Zhang,
Shengwu~Xiong,
Henan~Wang,
Wenjie~Yin,
Jiawang~Peng,
Duan~Zhou,
Yuqiang~Zhang,
Chen~Zhou,
Hua~Chen,
Qile~Zhao,
and~Pengfei~Duan
\thanks{Yifang~Zhang, Pengfei~Duan are with the Sanya Science and Education Innovation Park, Wuhan University of Technology, Sanya, 572000, China and also with the School of Computer Science and Artificial Intelligence, Wuhan University of Technology, Wuhan 430070, China (e-mail: yifangzhang@whut.edu.cn;
duanpf@whut.edu.cn).}%
\thanks{Shengwu~Xiong, Henan~Wang, Jiawang~Peng, and Duan~Zhou are with the School of Computer Science and Artificial Intelligence, Wuhan University of Technology, Wuhan 430070, China (e-mail: 
xiongsw@whut.edu.cn;
361332@whut.edu.cn; 
297975@whut.edu.cn;
zhouduan@whut.edu.cn).}%
\thanks{Wenjie~Yin, Yuqiang~Zhang, and Chen~Zhou are with the School of Earth and Space Science and Technology, Wuhan University, Wuhan 430072, China (e-mail: windsoryin@whu.edu.cn;
yqzhang\_3@whu.edu.cn;
chenzhou@whu.edu.cn).}%
\thanks{Hua~Chen is with the School of Water Resources and Hydropower Engineering, Wuhan University, Wuhan 430062, China (e-mail: chua@whu.edu.cn).}%
\thanks{Qile~Zhao is with the GNSS Research Center, Wuhan University, Wuhan 430062, China (e-mail: zhaoql@whu.edu.cn).}%
\thanks{Corresponding authors: Pengfei Duan (duanpf@whut.edu.cn).}

}

\markboth{Journal of \LaTeX\ Class Files,~Vol.~14, No.~8, August~2021}%
{Shell \MakeLowercase{\textit{et al.}}: A Sample Article Using IEEEtran.cls for IEEE Journals}


\maketitle

\begin{abstract}
Global navigation satellite systems (GNSS) station-based Precipitation Nowcasting aims to predict rainfall within the next 0–6 hours by leveraging a GNSS station’s historical observations of precipitation, GNSS-PWV, and related meteorological variables, which is crucial for disaster mitigation and real-time decision-making. In recent years, time-series forecasting approaches have been extensively applied to GNSS station-based precipitation nowcasting. However, the highly imbalanced temporal distribution of precipitation, marked not only by the dominance of non-rainfall events but also by the scarcity of extreme precipitation samples, significantly limits model performance in practical applications. To address the dual imbalance problem in precipitation nowcasting, we propose a continuous probability modeling–based framework, \textbf{\textit{RainBalance}}. This plug-and-play module performs clustering for each input sample to obtain its cluster probability distribution, which is further mapped into a continuous latent space via a variational autoencoder (VAE). By learning in this continuous probabilistic space, the task is reformulated from fitting single and imbalance-prone precipitation labels to modeling continuous probabilistic label distributions, thereby alleviating the imbalance issue. We integrate this module into multiple state-of-the-art models and observe consistent performance gains. Comprehensive statistical analysis and ablation studies further validate the effectiveness of our approach. 
\end{abstract}


\section{Introduction}
Accurately predicting rainfall within the next 0–6 hours, commonly referred to as precipitation nowcasting, is vital for effective disaster mitigation, flood management, and real-time operational decision-making \cite{nearing2024global}. Traditional physics-based numerical weather prediction (NWP) models, based on differential equations, are primarily suited for medium-term forecasting. 
For nowcasting, data-driven methods predominate, including weather radar-based rainfall extrapolation \cite{ravuri2021skilful,zhang2023skilful} (e.g., DGMR and NowcastNet), satellite-based rainfall prediction \cite{rahimi2024global,shukla2025satellite}, and GNSS station-based time series forecasting using meteorological data \cite{li2022improved,yin2025accurate,an2025deep}. Among these approaches, GNSS-based methods have recently gained increasing attention due to their ability to observe atmospheric water vapor processes that are directly linked to precipitation formation.

GNSS signals experience tropospheric delay as they pass through the atmosphere. The tropospheric delay mainly refers to zenith total delay (ZTD), which consists of zenith hydrostatic delay (ZHD) and zenith wet delay (ZWD) \cite{zhao2020improved}. The ZWD can be converted into precipitable water vapor (PWV) \cite{bevis1994gps}, which provides high-resolution and all-weather measurements of atmospheric moisture. Numerous studies have demonstrated that rapid increases or anomalous peaks in PWV often precede the onset of rainfall or extreme precipitation events \cite{yao2017establishing,gong2023assimilating,wu2021high}. This strong and well-understood physical relationship offers a solid foundation for leveraging GNSS-derived atmospheric parameters in predictive modeling.

Consequently, GNSS station-based precipitation nowcasting utilizes a GNSS station’s historical time series of precipitation, PWV, and auxiliary meteorological variables to independently predict future rainfall at that specific location.
This approach is particularly applied in providing highly customized forecasting services for key sites, such as urban waterlogging hotspots and critical hydraulic facilities. Typically, this task is formulated as a time series forecasting problem.

However, accurate GNSS station-based precipitation nowcasting faces its own distinct challenge: the dual imbalance problem, which is the central focus of this work.
\begin{figure}[!t]
\centering
\includegraphics[width=0.9\linewidth]{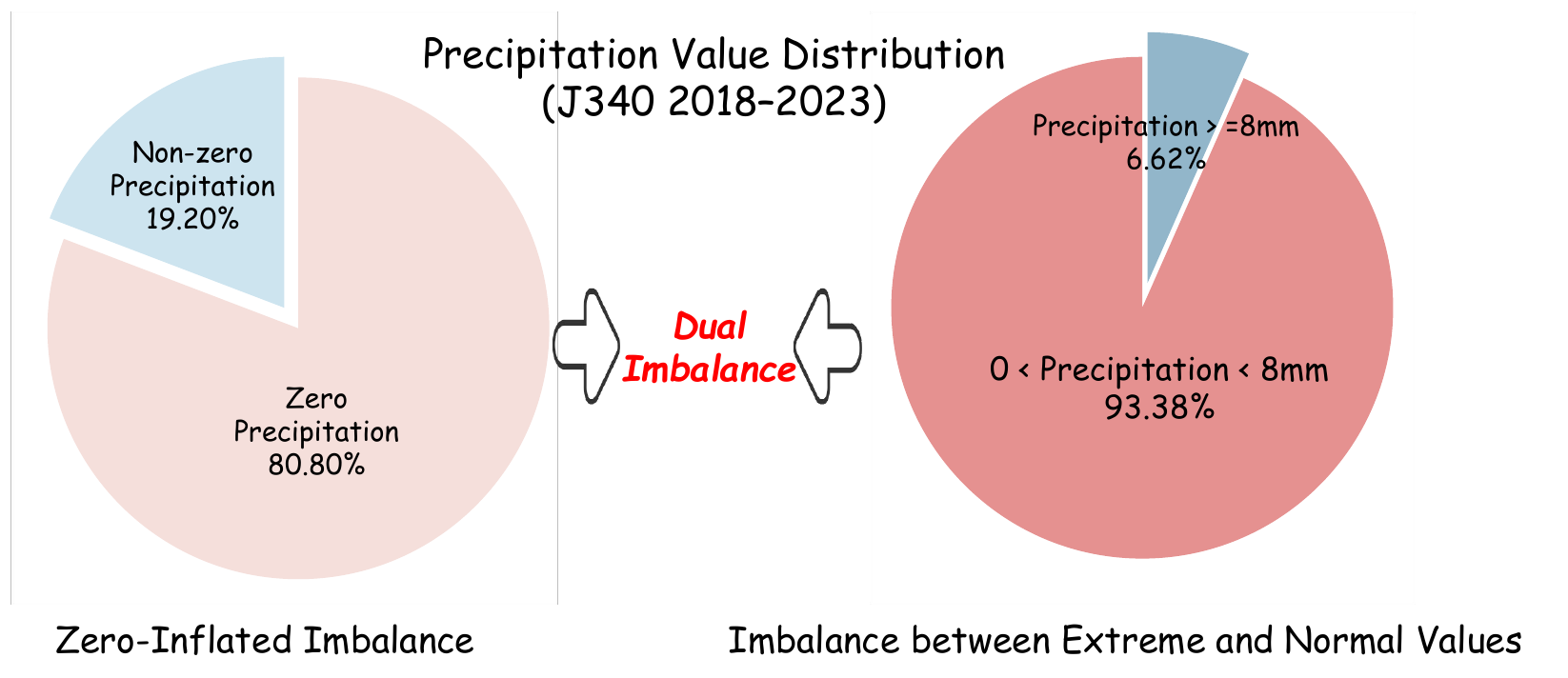}
\caption{Distribution of precipitation values at the J340 GNSS station. The left panel shows the proportion of zero and non-zero precipitation samples, revealing a strong zero-inflation (80.80\% of all samples record no rainfall). The right panel further decomposes the non-zero samples, indicating that light rainfall events (0–8 mm/h) dominate the dataset (93.38\%), while extreme precipitation events (>=8 mm/h) are relatively rare (6.62\%).} 
\label{fig_tp_distribution}
\end{figure}
Specifically, this dual-imbalance arises from the sparsity of rainy events and the skewed intensity distribution among them. It can be clearly illustrated in Fig.~\ref{fig_tp_distribution}.
Based on six years of precipitation time series from a single station, over 80\% of the time steps contain no rainfall, while only about 1.27\% correspond to extreme events. 

Although data imbalance is a shared challenge across many time series domains, the problem in precipitation forecasting is unique. This contrasts with domains such as stock market or traffic flow data, where imbalance is primarily one-dimensional and arises mainly from variations in activity levels over time \cite{li2023extreme,zhang2024irregular}. For example, in stock market data, fluctuations are usually small, with significant volatility occurring only during specific events (e.g., financial crises or corporate earnings announcements)\cite{wang2025pre}. Similarly, traffic flow data shows large volumes during peak hours (e.g., rush hours) and very low volumes during nighttime or holidays, reflecting imbalanced distributions across different time intervals\cite{huang2024multivariate}. While these domains also exhibit imbalance, it mainly arises from variations in magnitude over time, rather than the dual-imbalance of zero dominance and extreme scarcity observed in precipitation data. Therefore, the dual-imbalance presents distinct challenges for traditional time-series models.
Since deep learning networks are prone to underestimating extreme events due to significant data imbalance, the model's performance is often compromised, particularly when forecasting rare but impactful occurrences such as heavy rainfall. The coexistence of these two forms of imbalance further amplifies model bias.

\begin{figure}[h]
\centering
\includegraphics[width=1.0\linewidth]{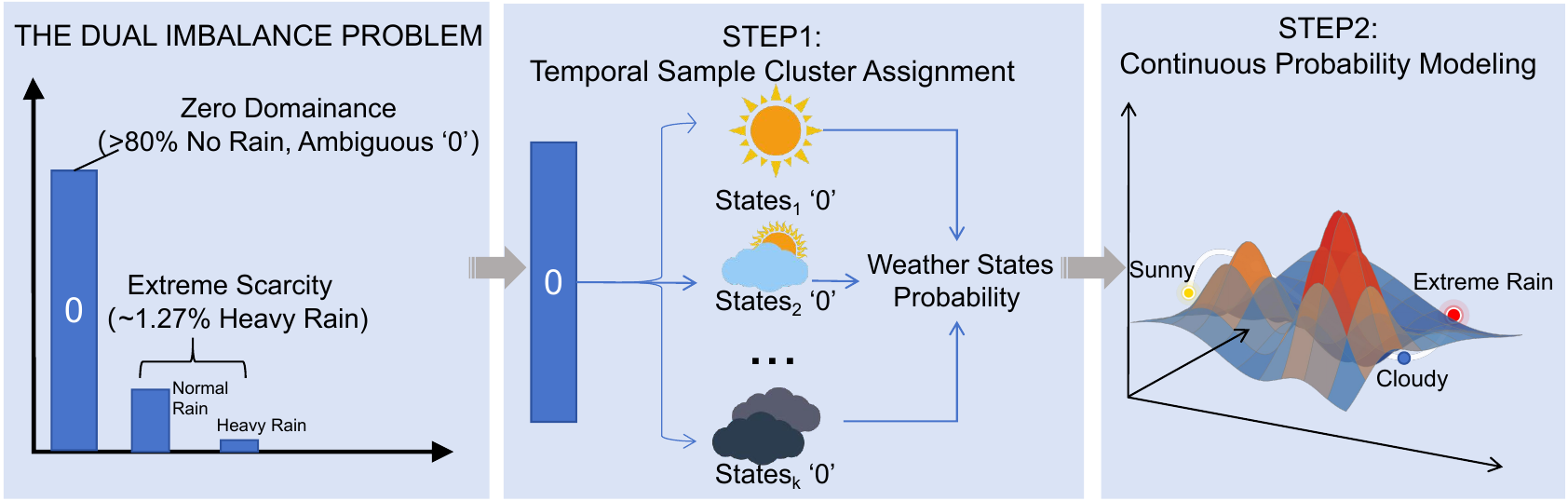}
\caption{An Overview of RainBalance for Addressing the Dual Imbalance Problem in Precipitation Nowcasting. This diagram illustrates a staged methodology for tackling both zero-dominance and extreme scarcity in precipitation data. It begins by clustering ambiguous "no-rain" states into multi weather states, and then proceeds with continuous probability modeling to effectively capture and forecast precipitation events.} 
\label{fig_motivation}
\end{figure}

Existing approaches to handle imbalance, such as data resampling or cost-sensitive learning, primarily tackle the sample imbalance at a superficial level \cite{cerqueira2024time,li2024learning,silvestrin2021framework}. They do not address the core issue of imbalance:  
a single precipitation label may correspond to a continuum of distinct atmospheric states, even when the label value is the same.
Specifically,
a single "zero rainfall" label can correspond to a multitude of distinct atmospheric states (e.g., clear sky vs. pre-storm humidity), causing the model to overlook subtle pre-precipitation signals.
There is a pressing need for bridging deterministic mapping with probabilistic modeling to respect the continuous and uncertain dynamics of weather systems.

To fundamentally resolve this, we propose \textbf{RainBalance}, a continuous-space probabilistic reformulation tailored to dual-imbalanced time series, as shown in Fig.~\ref{fig_motivation}. Our core insight is to transition from fitting single, imbalanced labels to modeling continuous, probabilistic label distributions. 
This is achieved through a two-stage process: First, for each input sequence, we perform sample-wise clustering to obtain a discrete probability distribution over learned "weather prototypes." This step elevates raw data to a more conceptual level. Second, to embody the continuous essence of meteorological processes, we employ a VAE\cite{kingma2019introduction} to map this discrete distribution into a smooth, continuous latent space. This continuous evolution space allows for seamless interpolation between different weather states, representing the gradual transitions observed in nature.

By operating within this learned continuous space, RainBalance enriches the model's representations with a continuous probabilistic prior. This guides the model to refine its internal representations, sharpening its ability to differentiate latent signatures of rare events.

We applied our proposed module to various state-of-the-art (SOTA) time series forecasting models and validated it on precipitation data collected from different regions. The results demonstrate that our \textbf{RainBalance} module significantly improves the performance of existing time series models and exhibits strong generalizability. In summary, the contributions of this paper are as follows:
\begin{itemize}
\item
\textbf{Definition of the Dual-Imbalance problem:} We formally define a time series forecasting problem characterized by dual imbalance—dominated by excessive zero values and scarce extreme events. Unlike traditional imbalance caused by varying magnitudes, this dual imbalance makes accurate forecasting more difficult.

\item
\textbf{Proposed RainBalance module:} We propose \textbf{RainBalance}, a plug-and-play module that mitigates dual imbalance by mapping discrete target labels into a continuous probabilistic space through clustering and a VAE. This enhances prediction accuracy for rare and extreme events.

\item
\textbf{Convincing results}: We integrate RainBalance into multiple state-of-the-art forecasting models and evaluate it on global precipitation datasets. The results consistently show significant improvements, demonstrating its robustness and generalizability.
\end{itemize}

\section{Related Work}

\subsection{Time-series Model for Precipitation forecasting}

Harilal et al.\cite{THOTTUNGALHARILAL2024108581}, using meteorological data from 1981 to 2023 across four regions in the UK, employed hybrid deep learning models (CNN-LSTM and RNN-LSTM) for daily precipitation prediction. The experimental results demonstrated that these hybrid models outperformed traditional LSTM and its variants in performance. Yin et al.\cite{yin2025accurate}, based on the Informer\cite{zhou2021informer} model, integrated GNSS PWV data and ERA5 meteorological data, effectively capturing long-range dependencies in time series data. Geng et al.\cite{geng2023lstmatu} introduced the LSTMAtU-Net model, which combines LSTM units with a U-Net structure and incorporates Efficient Channel and Spatial Attention (ECSA) modules. This integration significantly enhanced the model's ability to capture long-term dependencies and spatial features, improving performance across various precipitation thresholds in precipitation nowcasting.

Existing time-series models for precipitation forecasting primarily emphasize variable interactions and spatiotemporal dependencies, but seldom account for the distinctive characteristics of precipitation time-series data.

\subsection{Imbalance Problem in Time-series Forecasting}

Time series forecasting with highly imbalanced labels has been extensively studied, and existing methods can be broadly categorized into three classes. First, \textbf{data-level strategies}, including oversampling of rare events or synthetic data generation, aim to increase the presence of infrequent extreme values \cite{cerqueira2024time,li2024learning,silvestrin2021framework}. While effective in exposing the model to rare cases, these approaches may introduce distributional bias and distort the temporal dynamics inherent in the original sequences. Second, \textbf{model-level strategies}, such as hybrid distribution models or multi-branch networks, explicitly separate the prediction of frequent normal values and rare extremes\cite{shi2024boosting,wang2023generalized}. These methods can improve tail prediction but often rely on manually defined thresholds or distributional assumptions, limiting their generality. Third, \textbf{loss-level strategies}, including reweighted or extreme-aware objectives, emphasize rare events during training, yet they are sensitive to hyperparameters and may lead to unstable optimization when the imbalance is severe\cite{chaomin2025contrastive,wang2024self}. A common thread among these approaches is their tendency to treat extreme events discretely, either by amplifying rare labels or separating them from the main prediction task, which underutilizes the continuous structure of the label space.

In precipitation forecasting, the problem is more severe: \textbf{dual imbalance} exists, where \textit{no-rain vs. rain} imbalance overlaps with \textit{normal vs. extreme} value imbalance. This compounded imbalance has rarely been addressed in prior research, yet it is a defining characteristic of meteorological data, especially in precipitation forecasting. Tackling this dual imbalance is therefore crucial for building robust and practically useful forecasting models.

\subsection{Cluster Method in in Time-series Forecasting}
Chen et al. \cite{chen2024similarity} proposed the Channel Clustering Module (CCM), which dynamically groups input channels based on intrinsic similarity and learns cluster prototypes via a cross-attention mechanism. CCM aims to balance Channel-Independent and Channel-Dependent modeling strategies and enhance general multivariate time-series forecasting.
Qiu et al. \cite{qiu2025duet} proposed DUET, a general multivariate time series forecasting framework that simultaneously performs clustering along both temporal and channel dimensions. DUET's Temporal Clustering Module (TCM) clusters the temporal distribution patterns of each channel, assigning different encoders to channels that exhibit heterogeneous temporal dynamics. In essence, TCM performs pattern-aware embedding for each channel’s time series.

DUET’s TCM and Chen et al.’s CCM both perform clustering at the channel level, aiming to enhance channel representations. In contrast, RainBalance clusters individual temporal samples within each sequence to construct a continuous probabilistic latent space, redefining supervision signals to directly tackle the dual-imbalance problem. Unlike TCM and CCM, which focus on feature-level representations, RainBalance operates at the label modeling level, providing task-specific probabilistic supervision tailored for precipitation nowcasting.

\section{Preliminary}
\subsection{Problem Definition}
We formulate station-based precipitation nowcasting as a multivariate-to-univariate time series prediction task. Given a sequence of historical observations comprising both meteorological variables and past precipitation values, the goal is to predict future precipitation over a fixed horizon.

Formally, let the input sequence be defined as:
\begin{equation}
\begin{aligned}
\mathbf{X} = \{ \mathbf{x}_i, {y}_{i} \}_{t=i}^{i+l-1}, \quad \mathbf{x}_i \in \mathbb{R}^D
\end{aligned}
\end{equation}
where $\mathbf{x}_i$ includes meteorological factors (e.g., temperature, humidity, wind) and $\mathbf{y}_i$ is the precipitation at time $i$, and $D$ denotes the number of input variables. $l$ is the input length.

The target is to predict future rainfall:
\begin{equation}
\begin{aligned}
\mathbf{y} = \{ y_{i+l}, y_{i+l+1}, \dots, y_{i+l+h-1} \}, \quad y_{t} \in \mathbb{R}
\end{aligned}
\end{equation}
where $h$ is the prediction horizon. Notably, the output is univariate, focusing solely on future precipitation, despite the multivariate nature of the inputs.

This setting captures the practical demands of real-world precipitation nowcasting, where complex environmental factors are used to infer a single but highly critical target variable.

\subsection{Imbalance in Time-Series Data}

Dual Imbalance problem in time-series data can manifest in two main forms: \textbf{Imbalance between extreme and normal values}, and \textbf{Zero-inflation Imbalance}, which is a special case of imbalance. 

\textbf{1) Imbalance between Extreme and Normal Values:}
   Extreme values represent rare, high-impact events such as heavy rainfall, while normal values represent the majority of regular observations. The imbalance between these two classes of values can be mathematically represented as:
\begin{equation}
\begin{aligned}
   \mathbf{y} = \{ y_t \}_{t=1}^{T}, \quad y_t \in \mathbb{R}, \\ \quad P(y_t \text{ is extreme}) \ll P(y_t \text{ is normal})
\end{aligned}
\end{equation}
   Extreme events, although infrequent, are critical for predictions due to their higher impact.

\textbf{2) Zero-Inflation Imbalance:}  

Zero-inflation refers to time-series data in which the majority of values are zero (e.g., no rainfall), while only a small fraction of values are non-zero (e.g., rainfall events). This can be formally expressed as:

\begin{equation}
\begin{aligned}
\mathbf{y} = \{ y_t \}_{t=1}^{T}, \quad y_t \in \mathbb{R}, \quad P(y_t = 0) \gg P(y_t \neq 0)
\end{aligned}
\end{equation}

where \( P(y_t = 0) \) is the probability of observing a zero value at time \( t \), and \( P(y_t \neq 0) \) is the probability of observing a non-zero value.
In zero-inflated time-series, the model must focus on detecting rare non-zero events amidst the abundant zeros.

\textbf{3) Dual Imbalance:}
   Zero-inflation causes a \textbf{dual imbalance} problem, where both zero and non-zero values are imbalanced. First, the sequence is dominated by zeros (no rainfall event), and second, the non-zero values (when rainfall occurs) often exhibit further imbalance, where small non-zero values (e.g., light rainfall) occur far more frequently than extreme values (e.g., heavy rainfall). This dual imbalance can be formally represented as:

\begin{equation}
\underbrace{P(y_t = 0)}_{\text{Zero vs Non-zero}} \gg 
\underbrace{P(y_t^+ \le y_{\text{th}} \mid y_t > 0)}_{\text{Normal vs Extreme within rain}} \gg 
\underbrace{P(y_t^+ > y_{\text{th}} \mid y_t > 0)}_{\text{Extreme Rain}}
\end{equation}

   where $y_{\text{th}}$ is a threshold that marks extreme events. Hence, zero-inflation leads to both the overabundance of zero values and the imbalance among non-zero values.

The issue we focus on is the most complex imbalance problem in time-series data — \textbf{Dual Imbalance}.  In this context, how to improve the overall predictive performance of models when dual imbalance exists in time-series data is the central issue we aim to address.

\section{Method}
\begin{figure*}[!t]
\centering
\includegraphics[width=0.7\textwidth]{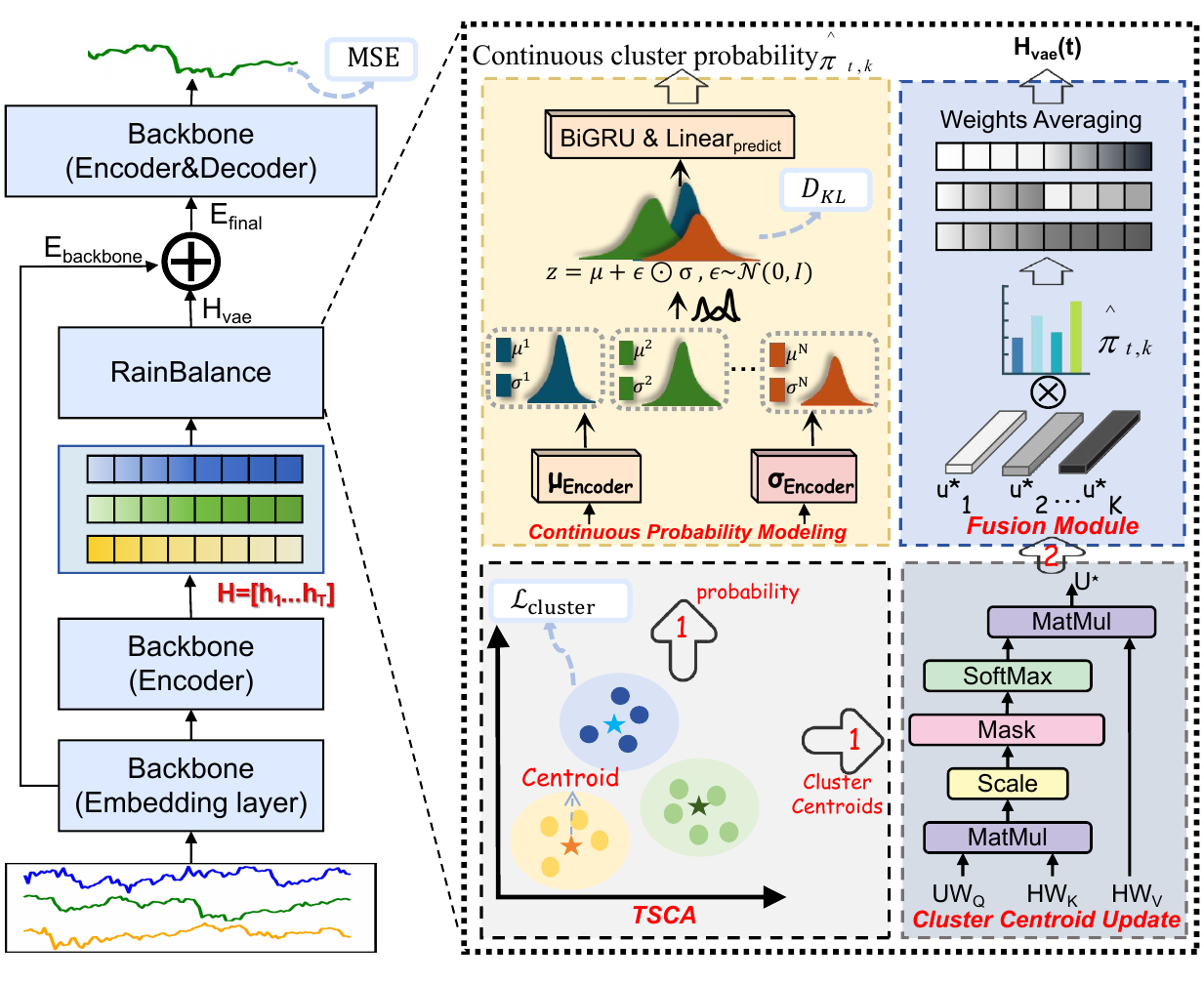}
\caption{An overview of the proposed RainBalance framework. The input time series is processed by the encoder of a backbone model to obtain hidden embeddings. The Temporal Sample Cluster Assignment Module (lower left) computes sample similarities and assigns samples to clusters, generating a cluster probability distribution and prototype embeddings. The Continuous Probability Modeling Module (upper left), structured as a Variational Autoencoder (VAE), encodes this distribution into a latent space z and decodes it to obtain new cluster probabilities. Finally, the Fusion Prediction Module (upper right) integrates the backbone embeddings with the weighted cluster prototypes to produce the final, balanced precipitation forecast.} 
\label{fig_intro}
\end{figure*}

We propose \textit{RainBalance}, a plug-and-play module that can be integrated into various time-series forecasting backbones to handle the dual imbalance in precipitation nowcasting. As shown in Fig.~\ref{fig_intro}, RainBalance consists of three components: \textbf{1) The Temporal Sample Cluster Assignment Module} abstracts the input sequence into discrete meteorological prototypes. \textbf{2) The Continuous Probability Modeling Module} infers a continuous probability distribution across these prototypes. \textbf{3) The Fusion Prediction Module} strategically conditions the backbone's representations on this distribution, guiding the model toward more balanced and probabilistic forecasts.

These modules work jointly with the base model to enhance its ability to model diverse precipitation regimes while keeping the overall structure lightweight and compatible with existing frameworks.

\subsection{Temporal Sample Cluster Assignment}
Inspired by the Cluster Assignment mechanism of CCM \cite{chen2024similarity}, we adapt it to perform clustering over individual temporal samples at each time step, rather than across channels, and refer to this adapted mechanism as Temporal Sample Cluster Assignment (TSCA).
\subsubsection{Sample Similarity}
Let \( \mathbf{X} = [\mathbf{x}_1, \mathbf{x}_2, \dots, \mathbf{x}_T] \) be a batch of samples, where each sample \( \mathbf{x}_i \in \mathbb{R}^C \).

\textbf{Compute squared Euclidean distance:}
The squared Euclidean distance between \( \mathbf{x}_i \) and \( \mathbf{x}_j \) is given by:
\begin{equation}
\begin{aligned}
\text{dist\_squared}_{ij} = \sum_{k=1}^{C} (\mathbf{x}_{ik} - \mathbf{x}_{jk})^2
\end{aligned}
\end{equation}
where \( \mathbf{x}_{ik} \) and \( \mathbf{x}_{jk} \) are the \( k \)-th elements of samples \( \mathbf{x}_i \) and \( \mathbf{x}_j \).

\textbf{Normalize the distance and compute similarity:}
Let \( \text{param} \) be the maximum of the squared distances:
\begin{equation}
\begin{aligned}
\text{param} = \max(\text{dist\_squared})
\end{aligned}
\end{equation}
Then, compute the similarity using the exponential function:
\begin{equation}
\begin{aligned}
\text{euc\_similarity}_{ij} = \exp\left(-\sigma \cdot \frac{\text{dist\_squared}_{ij}}{\text{param}}\right)
\end{aligned}
\end{equation}
where $\sigma$ is the hyper parameter controlling the spread of the Gaussian similarity function.

\textbf{Construct the similarity matrix:}
The similarity matrix \( \mathbf{S} \) contains the pairwise similarities between all samples:
\begin{equation}
\begin{aligned}
\mathbf{S} = \left[ \text{euc\_similarity}_{ij} \right]^l_{i,j=1}
\label{compute:S}
\end{aligned}
\end{equation}

\subsubsection{Cluster Assignment}
To discover latent temporal patterns within the sequence, we introduce 
\(K\) trainable cluster centroids 
\(
\mathcal{U}=\{\mathbf{u}_1,\ldots,\mathbf{u}_K\},
\mathbf{u}_k\in\mathbb{R}^{d},
\)
where \(d\) denotes the hidden dimension.  
Given an input series 
\(
X\in\mathbb{R}^{T\times C},
\)
the encoder produces a sequence of timestamp representations
\(
H=[\mathbf{h}_1,\ldots,\mathbf{h}_T]^\top\in\mathbb{R}^{T\times d}.
\)

Each timestamp is softly assigned to the centroids through cosine-based affinities:
\begin{equation}
\pi_{t,k}
=
\frac{
\exp\!\left(
\frac{\mathbf{u}_k^\top\mathbf{h}_t}{\|\mathbf{u}_k\|\|\mathbf{h}_t\|}
\right)
}{
\sum_{j=1}^{K}
\exp\!\left(
\frac{\mathbf{u}_j^\top\mathbf{h}_t}{\|\mathbf{u}_j\|\|\mathbf{h}_t\|}
\right)
},
\qquad
\sum_k\pi_{t,k}=1
\end{equation}

To obtain discrete yet differentiable cluster indicators, we adopt a Bernoulli-style stochastic relaxation:
\begin{equation}
A_{t,k}\approx \mathrm{Bernoulli}(\pi_{t,k}), \quad
A\in\mathbb{R}^{T\times K}
\end{equation}
The matrix \(A\) encodes approximate memberships and will guide cluster centroids refinement.

\subsubsection{Cluster Centroid Update}
Cluster Centroids are further refined by selectively aggregating timestamp features.  
Let centroid and sample embeddings be projected into the query--key--value spaces via learnable matrices \(W_Q, W_K, W_V\).  
The initial attention logits are computed as
\begin{equation}
\Gamma = \frac{ (U W_Q)(H W_K)^\top }{\sqrt{d}},
\qquad
U=[\mathbf{u}_1,\ldots,\mathbf{u}_K]^\top
\end{equation}

To ensure that each centroid only incorporates information from its assigned samples, we impose the cluster mask:
\begin{equation}
\tilde{\Gamma}
=
\mathrm{Normalize}\!\left(
\exp(\Gamma)\odot A^\top
\right)
\end{equation}
where \( \odot \) denotes elementwise multiplication.  
The refined cluster centroids are thus obtained by
\begin{equation}
U^{\star}
=
\tilde{\Gamma}\,(HW_V),
\qquad U^{\star}\in\mathbb{R}^{K\times d}
\end{equation}

These updated cluster centroids for the next round of cluster assignment.

\subsubsection{Cluster Loss}
To encourage coherent and well-separated cluster structures, we define a self-supervised clustering objective.  

Inspired by the CCM \cite{chen2024similarity}, the cluster loss function is formulated as:
\begin{equation}
\mathcal{L}_{\mathrm{cluster}}
=
-\mathrm{Tr}(A^\top \mathbf{S} A)
+
\mathrm{Tr}\!\left((I - A A^\top) \mathbf{S}\right)
\end{equation}

The first term maximizes similarity within each cluster, promoting compactness.  
The second term reduces similarity across clusters, enhancing separability.
Together, these terms allow the model to learn meaningful prototypes in a fully self-supervised manner.

\subsection{Continuous Probability Modeling}
Inspired by Variational Autoencoders (VAE), we design two GRU-based encoders to adaptively capture the continuous distribution for cluster probability. The encoder architecture consists of a bidirectional GRU, formulated as:

\begin{equation}
\begin{aligned}
\mathbf{h_{\mu}} = \mu_{\text{Encoder}}(\pi_{t}) = \text{BiGRU}(\pi_{t,k})
\end{aligned}
\end{equation}
\begin{equation}
\begin{aligned}
\mathbf{h_{\sigma}} = \sigma_{\text{Encoder}}(\pi_{t}) = \text{BiGRU}(\pi_{t,k})
\end{aligned}
\end{equation}
where $\mathbf{h}_{\mu}, \mathbf{h_{\sigma}} \in \mathbb{R}^{2 \times \text{hidden\_dim}}$ represents the hidden state from the bidirectional GRU at time step $t$.

The variational posterior distribution parameters are then obtained through linear projections:

\begin{equation}
\begin{aligned}
\boldsymbol{\mu}^2 &= \mathbf{W}_{\mu} \cdot \mathbf{h}_{\mu} + \mathbf{b}_{\mu}
\end{aligned}
\end{equation}
\begin{equation}
\begin{aligned}
\log \boldsymbol{\sigma}^2 &= \mathbf{W}_{\sigma} \cdot \mathbf{h}_{\sigma} + \mathbf{b}_{\sigma}
\end{aligned}
\end{equation}
where $\boldsymbol{\mu^2}$ and $\log \boldsymbol{\sigma}^2$ represent the mean and log-variance of the approximate posterior distribution $q(z|P)$.

To enable differentiable sampling from the posterior distribution, we employ the reparameterization trick:

\begin{equation}
\begin{aligned}
\mathbf{z} = \boldsymbol{\mu} + \boldsymbol{\epsilon} \odot \exp(0.5 \cdot \log \boldsymbol{\sigma}^2), \quad \boldsymbol{\epsilon} \sim \mathcal{N}(\mathbf{0}, \mathbf{I})
\end{aligned}
\end{equation}

Subsequently, a GRU-based temporal model is used to predict the clustering probabilities $\hat{\pi}_{t,k}$ for the labels.

\begin{equation}
\begin{aligned}
\mathbf{h}_{\text{predict}} &= \text{BiGRU}(\mathbf{z})
\end{aligned}
\end{equation}
\begin{equation}
\begin{aligned}
\hat{\pi}_{t,k} &= \mathbf{W}_{\text{predict}} \cdot \mathbf{h}_{\text{predict}} + \mathbf{b}_{\text{predict}}
\end{aligned}
\end{equation}
where $\mathbf{h}_{\text{predict}} \in \mathbb{R}^{2 \times \text{hidden\_dim}}$ denotes the hidden state of the GRU cell. $\hat{\pi}_{t,k}$ represents the predicted probability distribution over $K$ clusters for the input time step $t$. 

The Kullback-Leibler (KL) divergence between the approximate posterior distribution \( q(z|X) \) and the prior distribution \( p(z) \) is defined as:

\begin{equation}
\begin{aligned}
D_{\text{KL}}(q(z|P) \parallel p(z)) = -\frac{1}{2} \sum_{n=1}^{N} \left( 1 + \log \sigma_n^2 - \mu_n^2 - \sigma_n^2 \right)
\end{aligned}
\end{equation}

where, \( N \) is the number of latent dimensions. The prior distribution \( p(z) \) is typically a standard normal distribution \( \mathcal{N}(0, 1) \), so \( \log p(z) = 0 \) under this assumption.

This KL divergence term encourages the learned latent distributions to stay close to the standard normal distribution, regularizing the VAE during training.

\subsection{Fusion Prediction Modules}

\subsubsection{Fusion Module}
Let the \textit{Temporal Sample Cluster Assignment} module provide the prototype representation \( \mathcal{U} \in \mathbb{R}^{K \times d} \), where \( K \) is the number of clusters and \( d \) is the dimensionality of the prototype space. Let the \textit{Continuous Probability Modeling} module output the clustering probabilities \( \hat\pi_{t,k} \in [0, 1] \) for each time point \( t \), representing the likelihood of sample \( t \) belonging to the \( k \)-th cluster.

The latent variable information \( H_{\text{vae}} \in \mathbb{R}^{b \times T \times d} \) for each time point is computed by taking the weighted sum of the cluster prototypes, weighted by the clustering probabilities \( \hat{\pi}_{t,k} \):
\begin{equation}
\begin{aligned}
    H_{\text{vae}}(t) = \sum_{k=1}^{K} \hat{\pi}_{t,k} \cdot \mathcal{U}(k)
\end{aligned}
\end{equation}
where \( \mathcal{U}(k) \) is the cluster centroid for the \( k \)-th cluster. \( b \) is the batch size. \( T \) is the sequence length and \( d \) is the dimensionality of the latent space.

Let the embedding representation from the Backbone model be \( \mathbf{E}_\text{backbone} \in \mathbb{R}^{b \times T \times d} \). The final integrated representation is obtained by element-wise addition of the latent variable information \( H_{\text{vae}} \) and the backbone embedding:
\begin{equation}
\begin{aligned}
    \mathbf{E}_\text{final} = \mathbf{E}_\text{backbone} + H_{\text{vae}}
\end{aligned}
\end{equation}
    This combined vector \( \mathbf{E}_\text{final} \) is then passed through the encoder and decoder of the backbone model, which learns from this integrated representation to produce the final output.

\subsubsection{Integrate into Backbone}

The method we propose is a plug-and-play approach that can be seamlessly integrated into any existing time-series model. The only requirement is to input the latent variables obtained from the previous steps into the model, which will then yield the final prediction results.

Let the latent variables \( \mathbf{H}_{\text{final}} \in \mathbb{R}^{b \times T \times d} \) be the output obtained from the previous steps. These latent variables are then passed to the backbone model, which can be any pre-existing time-series model \( f_\text{backbone} \).

The final prediction \( \hat{\mathbf{y}} \) is computed as:
\begin{equation}
\begin{aligned}
\hat{\mathbf{y}} = f_\text{backbone}(\mathbf{H}_{\text{final}})
\end{aligned}
\end{equation}
where \( \hat{\mathbf{y}} \in \mathbb{R}^{h \times 1} \) is the predicted output.

\subsection{Training Objective}
The training loss is composed of three terms: the clustering loss, the KL divergence loss from the VAE module, and the Mean Squared Error (MSE) loss between the predicted and true results. The total loss function is given by:

\begin{equation}
\begin{aligned}
\mathcal{L}_{\text{total}} = \beta \cdot \left( \mathcal{L}_\mathrm{cluster} + D_{\text{KL}} \right) + (1 - \beta) \cdot \text{MSE}(\mathbf{y}, \hat{\mathbf{y}})
\end{aligned}
\end{equation}
where, \( \mathcal{L}_C \) represents the clustering loss,
\( D_{\text{KL}}\) denotes the KL divergence loss between the approximate posterior \( q(z|X) \) and the prior \( p(z) \),
\( \text{MSE}(\mathbf{y}, \hat{\mathbf{y}}) \) is the mean squared error between the true values \( \mathbf{y} \) and the predicted values \( \hat{\mathbf{y}} \),
\( \beta \in [0, 1] \) is a weighting factor that controls the relative importance of the clustering and VAE terms versus the MSE loss.

\section{Experiments}
    \subsection{Experiment Setup}
All experiments are conducted on an NVIDIA H100 80GB GPU. To ensure the generalizability of the experiment, we selected four representative stations worldwide for the study. To ensure evaluation consistency, all results are computed using de-normalized actual rainfall. For robustness, each experiment is repeated three times, and the average performance is reported as the final result. The datasets were split into training, validation, and test sets in a 7:1:2 ratio. 

Under our experimental configuration, the number of clusters K and the loss weighting factor $\beta$ were set to 6 and 0.3, respectively. All other hyperparameters remained consistent with the optimal configurations specified in the original official implementations of each baseline model.

\begin{table*}[t]
\setlength\tabcolsep{2.5pt}
\centering
\caption{Performance comparison of base models with and without the proposed module on four representative stations with distinct temporal resolutions and locations: J340(34.406°N, 135.364°E, hourly), ZIMM (46.877°N, 7.465°E, hourly), P095 (39.698°N, -119.537°W, hourly), and JFNG (30.516°N, 114.491°E, 15-min). Values in `+Ours` that improve over the base model are shown in \textbf{bold}. Each row's best result is underlined. IMP(\%) indicates the percentage improvement of +Ours over the base model.}
\resizebox{0.97\textwidth}{!}{
\begin{tabular}{c|c|cccc|cccc|cccc|cccc|cc}
\toprule
\multirow{2}{*}{\textbf{Dataset}} &
\multirow{2}{*}{\textbf{Model}} &
\multicolumn{2}{c}{\textbf{xLSTM\cite{beck2025xlstm}}} & 
\multicolumn{2}{c|}{\textbf{+Ours}} &
\multicolumn{2}{c}{\textbf{TimeMixer\cite{wangtimemixer}}} & 
\multicolumn{2}{c|}{\textbf{+Ours}} &
\multicolumn{2}{c}{\textbf{TimeMixerPP\cite{wang2025timemixer++}}} & 
\multicolumn{2}{c|}{\textbf{+Ours}} &
\multicolumn{2}{c}{\textbf{TimesNet\cite{wutimesnet}}} &
\multicolumn{2}{c|}{\textbf{+Ours}} &
\multicolumn{2}{c}{\textbf{IMP(\%)}} \\ 

 & & MSE$\downarrow$ & MAE$\downarrow$ & MSE$\downarrow$ & MAE$\downarrow$ & MSE$\downarrow$ & MAE$\downarrow$ & MSE$\downarrow$ & MAE$\downarrow$ & MSE$\downarrow$ & MAE$\downarrow$ & MSE$\downarrow$ & MAE$\downarrow$ & MSE$\downarrow$ & MAE$\downarrow$ & 
 MSE$\downarrow$ & MAE$\downarrow$ & MSE$\uparrow$ & MAE$\uparrow$  \\ 
\midrule

\multirow{3}{*}{\rotatebox[origin=c]{90}{\textbf{J340}}} 
& 2  & 0.9425 & 0.2707 & \textbf{\underline{0.8989}} & \textbf{\underline{0.2598}} & 0.9927 & 0.2725 & \textbf{0.9803} & \textbf{0.2692} & 1.2929 & 0.3267 & \textbf{1.2235} & \textbf{0.3207} & 1.0165 & 0.2884 & 1.0357 & \textbf{0.2625} &  2.34 &4.02 \\
& 4  & 1.1526 & 0.3176 & \textbf{\underline{1.0008}} & \textbf{\underline{0.2904}} & 1.2483 & 0.3118 & \textbf{1.2402} & \textbf{0.2934} & 1.5541 & 0.3516 & \textbf{1.3545} & \textbf{0.3149} & 1.3257 & 0.2974 & \textbf{1.2813} & \textbf{0.2905} &  7.50 & 6.81 \\
& 6  & \underline{1.3087} & 0.3626 & 1.3402 & \textbf{\underline{0.3056}} & 1.4299 & 0.3380 & \textbf{1.3973} & \textbf{0.3202} & 1.5403 & 0.3653 & \textbf{1.4806} & \textbf{0.3475} & 1.4593 & 0.3342 & \textbf{1.4478} & \textbf{0.3119} &  1.13 & 8.13 \\
\midrule

\multirow{3}{*}{\rotatebox[origin=c]{90}{\textbf{ZIMM}}} 
& 2  & 1.7119 & 0.3194 & 1.7181 & \textbf{0.3082} & 1.6637 & 0.3131 & \textbf{\underline{1.6540}} & \textbf{\underline{0.3005}} & 2.0659 & 0.3823 & \textbf{1.8969} & \textbf{0.3543} & 1.6587 & 0.3198 & \textbf{1.6545} & \textbf{0.2979}  & 2.16 & 5.42 \\
& 4  & 1.9464 & 0.3457 & \textbf{\underline{1.8428}} & \textbf{\underline{0.3138}} & 1.8966 & 0.3419 & \textbf{1.8725} & \textbf{0.3411} & 2.1274 & 0.3597 & \textbf{2.0535} & 0.3698 & 1.8857 & 0.3426 & \textbf{1.8799} & \textbf{0.3330} & 2.59 & 2.36 \\
& 6  & 2.0317 & 0.3522 & \textbf{\underline{1.9711}} & \textbf{\underline{0.3171}} & 2.0095 & 0.3538 & \textbf{1.9899} & \textbf{0.3495} & 2.2055 & 0.3666 & \textbf{2.0022} & \textbf{0.3503} & 1.9884 & 0.3504 & \textbf{1.9829}&\textbf{0.3286} & 3.36 & 5.46 \\
\midrule

\multirow{3}{*}{\rotatebox[origin=c]{90}{\textbf{P095}}}
 & 2  & \underline{0.3359} & 0.0824 & 0.3417 & \textbf{\underline{0.0746}} & 0.3662 & 0.0892 & \textbf{0.3556} & \textbf{0.0750} & 0.3853 & 0.0948 & \textbf{0.3671} & \textbf{0.0904} & 0.3910 & 0.0952 & \textbf{0.3674} & \textbf{0.0857} & 2.98 & 10.05 \\
 & 4  & 0.3701 & 0.0899 & \textbf{0.3588} & \textbf{0.0879} & 0.3717 & 0.0911 & \textbf{0.3698} & \textbf{\underline{0.0812}} & 0.3887 & 0.0933 & \textbf{0.3850} & \textbf{0.0909} & 0.3786 & 0.0919 & \textbf{0.3715} & \textbf{0.0845} & 1.60 & 5.92 \\
 & 6  & 0.3864 & 0.0944 & \textbf{0.3653} & 0.0953 & 0.3865 & 0.0970 & \textbf{0.3773} & \textbf{\underline{0.0813}} & 0.3912 & 0.0915 & \textbf{0.3834} & 0.0954 & 0.3864 & 0.0942 & \textbf{0.3786} & \textbf{0.0898} & 2.97 & 3.89 \\
\midrule

\multirow{5}{*}{\rotatebox[origin=c]{90}{\textbf{JFNG}}}
 & 4  & 0.0191 & 0.0500 & \textbf{\underline{0.0145}} & \textbf{\underline{0.0253}} & 0.0175 & 0.0260 & \textbf{0.0157} & \textbf{0.0227} & 0.1730 & 0.1001 & 0.1736 & \textbf{0.0998} & 0.0180 & 0.0266 & \textbf{0.0167} & \textbf{0.0247} & 10.24 & 17.40 \\
 & 6  & 0.0372 & 0.0685 & \textbf{\underline{0.0283}} & \textbf{\underline{0.0361}} & 0.0367 & 0.0399 & \textbf{0.0323} & \textbf{0.0352} & 0.1856 & 0.1036 & 0.1862 & \textbf{0.1033} & 0.0473 & 0.0498& \textbf{0.0343} & \textbf{0.0372} & 15.80 & 21.18 \\
 & 8  & 0.0551 & 0.0830 & \textbf{\underline{0.0466}} & \textbf{\underline{0.0442}} & 0.0613 & 0.0536 & \textbf{0.0503} & \textbf{0.0459} & 0.1974 & 0.1065 & 0.1980 & \textbf{0.1063} & 0.0691 & 0.0597 & \textbf{0.0680} & \textbf{0.0580} & 8.70 & 16.05 \\
 & 10 & 0.0730 & 0.0914 & \textbf{\underline{0.0544}} & \textbf{\underline{0.0500}} & 0.0821 & 0.0641 & \textbf{0.0686} & \textbf{0.0557} & 0.2080 & 0.1092 & 0.2081 & \textbf{0.1091} & 0.0917 & 0.0710 & \textbf{0.0873} & \textbf{0.0683} & 11.66 & 15.57 \\
 & 12 & 0.0873 & 0.0958 & \textbf{\underline{0.0722}} & \textbf{\underline{0.0620}} & 0.0992 & 0.0709 & \textbf{0.0934} & \textbf{0.0659} & 0.2164 & 0.1121 & \textbf{0.1515} & \textbf{0.0805} & 0.1100 & 0.0779 & \textbf{0.1088} & 0.0785 & 13.54 & 17.43 \\
\bottomrule
\end{tabular}}
\label{tab:Ours_comparison}
\end{table*}

\subsection{Experiment Datasets}
The dataset we used is a GNSS-based Precipitation Nowcasting dataset, which includes 140 stations \cite{zhang2025effectivetimeseriesmodelsprecipitation,yin2025accurate} that are geographically distributed across different continents, elevations, and climate zones.

\begin{figure}[htbp]
    \centering
    \includegraphics[width=0.9\linewidth]{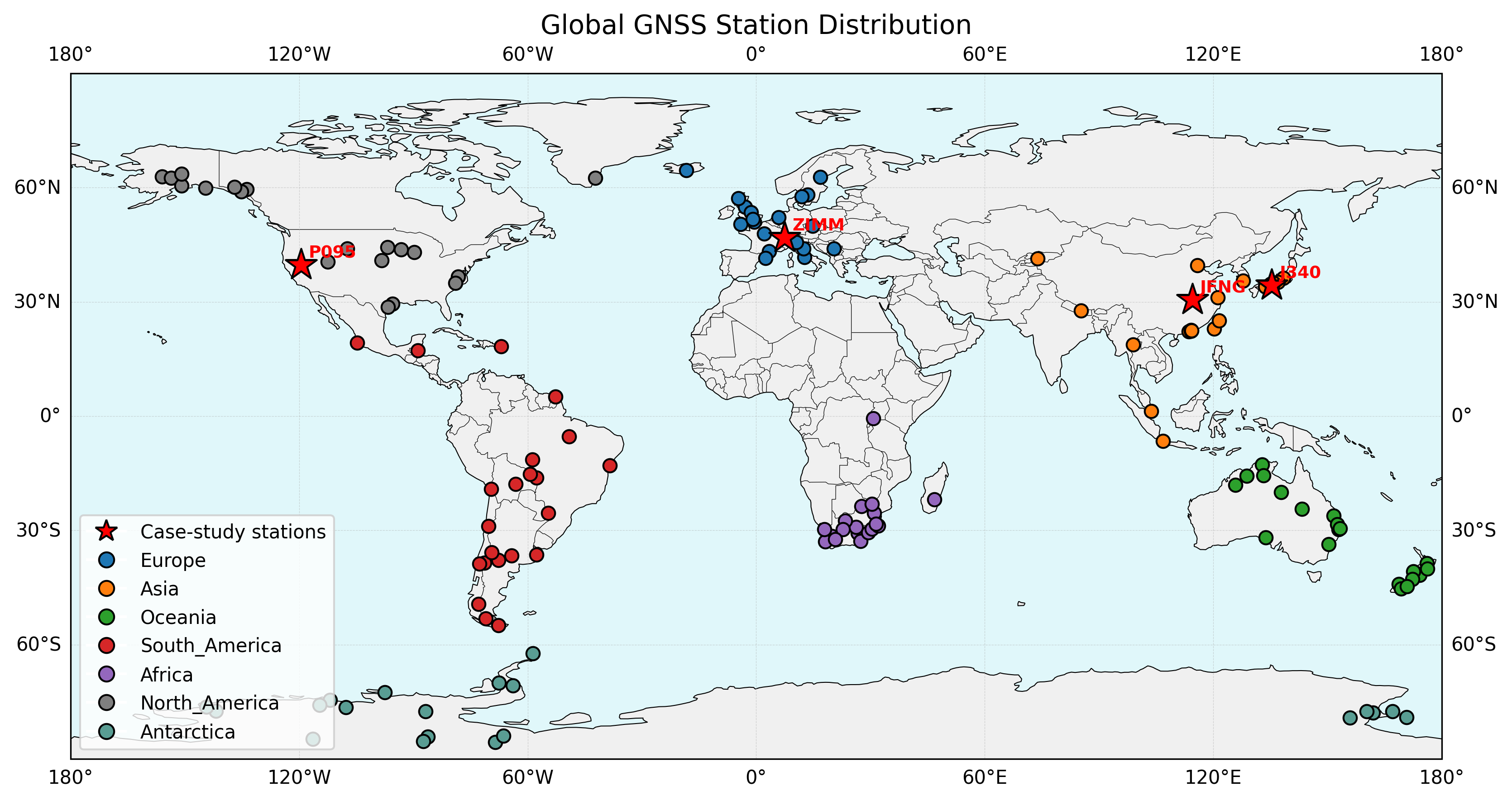}
    \caption{
      Global distribution of 140 selected GNSS stations from the RainfallBench dataset across seven continents, ensuring balanced spatial coverage for evaluating precipitation forecasting models \cite{zhang2025effectivetimeseriesmodelsprecipitation}. Four representative stations with different spatial distributions and temporal resolutions (J340, ZIMM, P095, and JFNG) were selected as case-study sites for detailed analysis.
    }
    \label{fig:station_distribution}
\end{figure}

This real-world meteorological dataset spans the period from January 1, 2018, 00:00 to January 1, 2024, 00:00, with observations recorded at hourly or 15-minute intervals. Each record consists of six variables (excluding the timestamp): five meteorological features and one target variable representing precipitation. Specifically, the input features include:
\begin{itemize}
    \setlength{\itemsep}{0.03em} 
    \item \textbf{t2m}: temperature at 2 meters above ground.
    \item \textbf{sp:} surface pressure
    \item \textbf{rh:} relative humidity
    \item \textbf{wind\_speed:} wind speed.
    \item \textbf{PWV:} precipitable water vapor, retrieved by inverting GNSS signal delays based on their proportional relationship with atmospheric water vapor.
    \item \textbf{tp:} total precipitation (target), obtained from GPM IMERG.
\end{itemize}

\subsection{Baseline Models}
To comprehensively evaluate the effectiveness of RainBalance, we organize the baselines into three categories: 
1) general-purpose time-series forecasting models, 
2) clustering-based representation learning methods, and 
3) imbalance-mitigation strategies. 
This design isolates the contributions of RainBalance from architectural capacity, clustering mechanisms, and loss-level imbalance handling.
\subsubsection{General Time-Series Forecasting Models}
RainBalance is a model-agnostic module designed to improve precipitation nowcasting performance. We carefully select four representative state-of-the-art models as base models: xLSTM\cite{beck2025xlstm}, TimeMixer\cite{wangtimemixer}, TimeMixerPP\cite{wang2025timemixer++}, and TimesNet\cite{wutimesnet}, which collectively cover four major paradigms, including RNN-based, MLP-based, Attention-based, and CNN-based architectures. For fair evaluation, we implement both the original base models and the RainBalance-enhanced versions using the optimal experimental configurations provided in their official repositories.

\subsubsection{Clustering-Based Representation Methods}

To distinguish RainBalance from existing clustering-driven modeling strategies, we include representative clustering-based baselines:
\begin{itemize}
    \item \textbf{Channel Clustering Module (CCM)}~\cite{chen2024similarity}: a channel-level clustering mechanism designed to improve temporal modeling efficiency.
\end{itemize}

\subsubsection{Imbalance-Mitigation Approaches}
We further evaluate RainBalance against methods that mitigate imbalance through tailored loss functions, including:
\begin{itemize}
    \item \textbf{Extreme Penalized Loss} (EPL) \cite{wang2024self}: the EPL is a piecewise loss function that employs MSE for normal values, while imposing an exponential, asymmetric penalty on extreme values exceeding a predefined threshold, thereby explicitly emphasizing the modeling of extreme events.
\end{itemize}

\begin{table}[t]
\setlength\tabcolsep{2.5pt}
\centering
\caption{Performance comparison of three representative enhancement modules—our proposed RainBalance module, the Channel Clustering Module (CCM), and the Extreme Penalized Loss (EPL)—integrated into the TimesNet. Each row’s best result is in \textbf{bold}. IMP(\%) denotes the percentage improvement achieved by our module over the TimesNet Model.}
\resizebox{0.97\linewidth}{!}{
\begin{tabular}{c|c|cccccccc|cc}
\toprule
\multirow{2}{*}{\textbf{Dataset}} &
\multirow{2}{*}{\textbf{Model}} &
\multicolumn{2}{c}{\textbf{TimesNet\cite{wutimesnet}}} &
\multicolumn{2}{c}{\textbf{+CCM\cite{chen2024similarity}}} &
\multicolumn{2}{c}{\textbf{+EPL\cite{wang2024self}}}&
\multicolumn{2}{c|}{\textbf{+Ours}} &
\multicolumn{2}{c}{\textbf{IMP(\%)}} \\ 

 & & MSE$\downarrow$ & MAE$\downarrow$ & MSE$\downarrow$ & MAE$\downarrow$ &
 MSE$\downarrow$ & MAE$\downarrow$ &
 MSE$\downarrow$ & MAE$\downarrow$ & MSE$\uparrow$ & MAE$\uparrow$  \\ 
\midrule

\multirow{3}{*}{\rotatebox[origin=c]{90}{\textbf{J340}}} 
& 2  & \textbf{1.0165} & 0.2884 &1.1395&0.3164&5.7028 &0.8127&1.0357 & \textbf{0.2625} &  -1.89 &8.98 \\
& 4  & 1.3257 & 0.2974 &1.3552&0.3470&5.3834&0.8193& \textbf{1.2813} & \textbf{0.2905} &  3.35 & 6.84 \\
& 6  & 1.4593 & 0.3342 &1.5355&0.3833&8.0321&0.9523& \textbf{1.4478} & \textbf{0.3119} &  0.79 & 6.67 \\
\midrule

\multirow{3}{*}{\rotatebox[origin=c]{90}{\textbf{ZIMM}}} 
& 2  &  1.6587 & 0.3198 &1.7059&0.3409&9.7433&1.0630& \textbf{1.6545} & \textbf{0.2979}  & 0.25 & 6.85\\
& 4  &  1.8857 & 0.3426 &2.0097&0.3770&9.3195&1.2133& \textbf{1.8799} & \textbf{0.3330} & 0.31 & 2.80 \\
& 6  &  1.9884 & 0.3504 &2.0933&0.3852&8.2290&1.1829&\textbf{1.9829}&\textbf{0.3286} & 0.28 & 6.22 \\
\midrule

\multirow{3}{*}{\rotatebox[origin=c]{90}{\textbf{P095}}}
 & 2  &  0.3910 & 0.0952 &0.3932&0.1162&1.1467&0.2725& \textbf{0.3674} & \textbf{0.0857} & 6.04 & 9.98 \\
 & 4  & 0.3786 & 0.0919 &0.3789&0.0973&1.4951&0.3586& \textbf{0.3715} & \textbf{0.0845} & 1.88 & 8.05 \\
 & 6  &  0.3864 & 0.0942 &0.3918&0.0981&1.4739&0.3390& \textbf{0.3786} & \textbf{0.0898} & 2.02 & 4.67\\
\midrule

\multirow{5}{*}{\rotatebox[origin=c]{90}{\textbf{JFNG}}}
 & 4  & 0.0180 & 0.0266 &0.0241&0.0342&0.3027&0.1123& \textbf{0.0167} & \textbf{0.0247} & 7.22 & 7.14 \\
 & 6  & 0.0473 & 0.0498&0.0639&0.0614&0.2803 &0.1099&\textbf{0.0343} & \textbf{0.0372} & 27.48 & 25.30 \\
 & 8  & 0.0691 & 0.0597 &0.0776&0.0656&0.2916 &0.1221&\textbf{0.0680} & \textbf{0.0580} & 1.59 & 2.85 \\
 & 10 & 0.0917 & 0.0710 &\textbf{0.0871}&0.0685&0.3507&0.1273& 0.0873 & \textbf{0.0683} & 4.80 & 3.80 \\
 & 12 & 0.1100 & 0.0779 &0.1091&\textbf{0.0758}&0.5905 &0.1712&\textbf{0.1088} & 0.0785 & 1.09 & -0.77 \\
\bottomrule
\end{tabular}}
\label{tab:Ours_comparison_CCM}
\end{table}
\subsection{Experiment Results}

Based on the experimental results presented in Table \ref{tab:Ours_comparison}, we report the mean squared error (MSE) and mean absolute error (MAE) across four real-world GNSS stations for comprehensive model evaluation. From the Table \ref{tab:Ours_comparison}, we observe that the proposed module consistently enhances the performance of base models across most experimental settings. Specifically, our method improves forecasting performance in 85.71\% of cases in MSE and 92.86\% of cases in MAE across 56 different experimental configurations. Remarkably, the proposed module achieves substantial improvements on xLSTM, with significant reductions in MSE by 25.48\%($0.0730\to0.0544)$ and MAE by 49.40\%($0.0500\to0.0253$) on the JFNG dataset. The IMP(\%) column of the table quantifies the average percentage improvement in terms of MSE/MAE, which demonstrates the consistent enhancement brought by our method across all datasets and prediction horizons. The experimental results indicate that our approach is particularly effective in scenarios with complex temporal patterns and diverse dataset characteristics, which commonly occur in real-world forecasting applications.

To further clarify the effectiveness of our method, we additionally compare RainBalance with two representative enhancement strategies: the clustering-based CCM module and the imbalance-oriented EPL loss function. The results show that RainBalance consistently surpasses CCM across all stations and prediction horizons, suggesting that channel-level clustering alone provides limited benefits when facing the dual imbalance inherent in GNSS-based precipitation data. In contrast, the EPL loss often leads to degraded performance. Because EPL applies an exponential penalty to underestimated extremes, its loss value grows excessively under the dual imbalance of our datasets, resulting in unstable gradients and even prediction divergence. Overall, RainBalance delivers the most stable and reliable improvements among all compared modules.

Then, our analysis of the experimental results is guided by the following five research questions:

\textbf{RQ1: }How does each component of the RainBalance module contribute to the overall performance?\textbf{\textit{(Ablation Studies)}}

\textbf{Generalization \& Robustness Analysis} include the the following four research questions:

\textbf{RQ2: }How dose the proposed RainBalance perform in each area? (\textbf{\textit{Analysis from the Dataset Perspective}})

\textbf{RQ3: }How does the proposed RainBalance perform performance vary with changes in predict length? (\textbf{\textit{Multi-temporal scale Evaluation}})

\textbf{RQ4: }How does the proposed RainBalance perform vary with changes in multi-forecast resolution? (\textbf{\textit{Multi-Forecast Resolution Evaluation}})

\textbf{RQ5: }How do the proposed RainBalance perform for extreme precipitation forecasting?  (\textbf{\textit{Extreme precipitation Evaluation}})

\subsection{Ablation Studies}

To quantitatively dissect the individual contribution of each key component within our proposed RainBalance module, we conduct a comprehensive ablation study. First, we replace the cluster-generated probability distribution with a uniform distribution to isolate the effect of structural representation learning. Second, we substitute the variational autoencoder with a simple linear layer to examine the necessity of probabilistic temporal modeling. By comparing the complete model against these two variants, we precisely quantify each component's contribution to resolving the dual-imbalance problem.
\begin{table}[h]
    \centering
    \caption{Ablation analysis of each module in RainBalance at the JFNG station. “24(4)” denotes an input sequence length of 24 and an output sequence length of 4; other notations follow the same convention. The \colorbox{red!25}{red} indicates the best-performing model.}
    \resizebox{\linewidth}{!}{
    \begin{tabular}{cc cc cc cc cc cc}
    \hline 
    \multicolumn{2}{c}{xLSTM} & \multicolumn{2}{c}{24(4)} & \multicolumn{2}{c}{24(8)} & \multicolumn{2}{c}{24(12)}\\
    \cline{1-2}
    Cluster& VAE &MSE$\downarrow$&MAE$\downarrow$&MSE$\downarrow$&MAE$\downarrow$&MSE$\downarrow$&MAE$\downarrow$\\
    \hline
    \ding{55} & \ding{55} & 0.0191&0.0500 & 0.0551&0.0830 & 0.0873&0.0958 \\
     \ding{55} & \checkmark & 0.0157&0.0311 & 0.0463&0.0579 & 0.0742&0.0641 \\
   \checkmark & \ding{55} & 0.0160&0.0290&\cellcolor{red!25}0.0458&0.0573&0.0762&0.0733\\
    \checkmark & \checkmark &\cellcolor{red!25}0.0145&\cellcolor{red!25}0.0253 &0.0466&\cellcolor{red!25}0.0442&\cellcolor{red!25}0.0722&\cellcolor{red!25}0.0620 \\
    \hline
    \end{tabular}
    }
    \label{tab_ablation_JFNG}
\end{table}

The ablation results in Table \ref{tab_ablation_JFNG} clearly validate the synergistic efficacy of the two core components in our proposed RainBalance module. When the cluster-based probability distribution is replaced with a uniform distribution (row 2), performance degrades notably across all horizons, underscoring the critical role of structural representation learning via sample clustering in capturing inherent sample heterogeneity. Similarly, substituting the variational autoencoder with a simple linear layer (row 3) leads to further deterioration, highlighting the indispensable contribution of probabilistic temporal modeling to maintain distributional continuity. The complete model (row 4) achieves the best results in nearly all cases, validating the complementary benefits of integrating both clustering-based structural representations and VAE-driven latent dynamics in mitigating the dual-imbalance problem.
\subsection{Generalization \& Robustness Analysis}
\subsubsection{Analysis from the Dataset Perspective}

We further evaluate the generalization capability of RainBalance across four GNSS stations (J340, ZIMM, P095, and JFNG) with distinct geographical and climatic characteristics. The results consistently indicate that RainBalance improves the forecasting performance of all baseline models on every regional dataset, while the extent of improvement varies with the inherent complexity of local precipitation patterns.
\begin{figure}[h]
\centering
\includegraphics[width=1.0\linewidth]{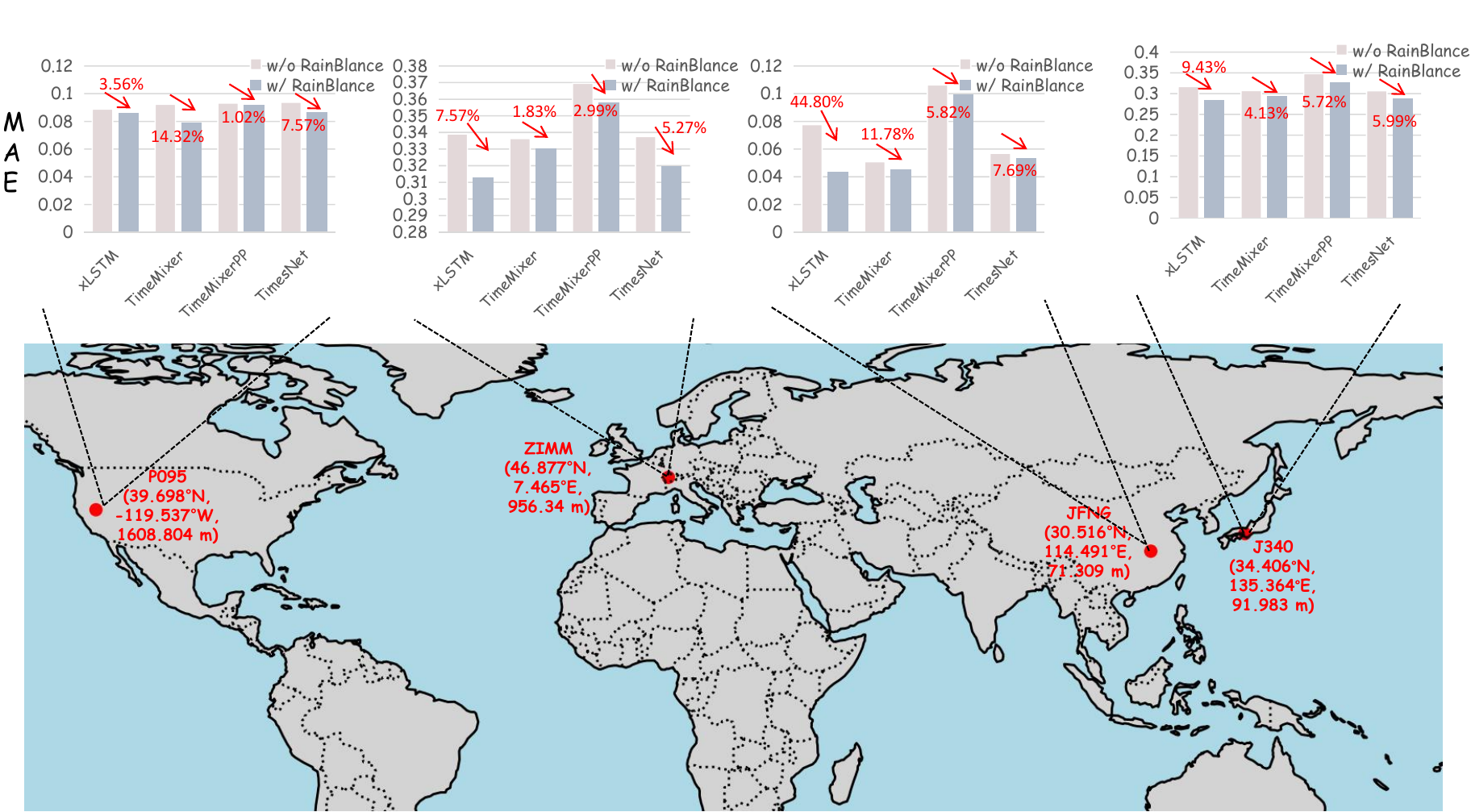}
\caption{Comparison of the Mean Absolute Error (MAE) improvement rates achieved by different configurations at various sites.} 
\label{fig_multi-scale}
\end{figure}

For regions with highly complex precipitation dynamics, such as the JFNG station located in the Yangtze River Basin of China (humid subtropical climate under strong East Asian monsoon influence), the improvement is most significant. The data from this region contain frequent zero-precipitation intervals and abrupt extreme rainfall events, resulting in a pronounced dual-imbalance problem. By projecting the discrete labels into a cluster-based continuous probability space, RainBalance effectively mitigates the learning bias and achieves substantial performance gains, with up to 25.48\%($0.0730\to0.0544)$ and 49.40\%($0.0500\to0.0253$) reductions in MSE and MAE, respectively.

In contrast, for stations in diverse but relatively stable climates—including J340 (humid subtropical, Japan), ZIMM (temperate continental, Switzerland), and P095 (arid desert, USA)—RainBalance still delivers consistent improvements across all base models. Specifically, it enhances the model’s sensitivity to rare rainfall events at the arid P095 site and improves the estimation of solid precipitation at the snow-dominated ZIMM site. These results suggest that RainBalance can effectively adapt to different climatic modes and precipitation regimes, providing robust gains without relying on region-specific patterns.
\subsubsection{Multi-temporal scale Evaluation}

We evaluated model performance under different prediction lengths (e.g., Model 2/4/6 for J340, P095, ZIMM sites and 4/6/8/10/12 for JFNG) to examine the temporal stability of RainBalance.
\begin{figure}[h]
\centering
\includegraphics[width=1.0\linewidth]{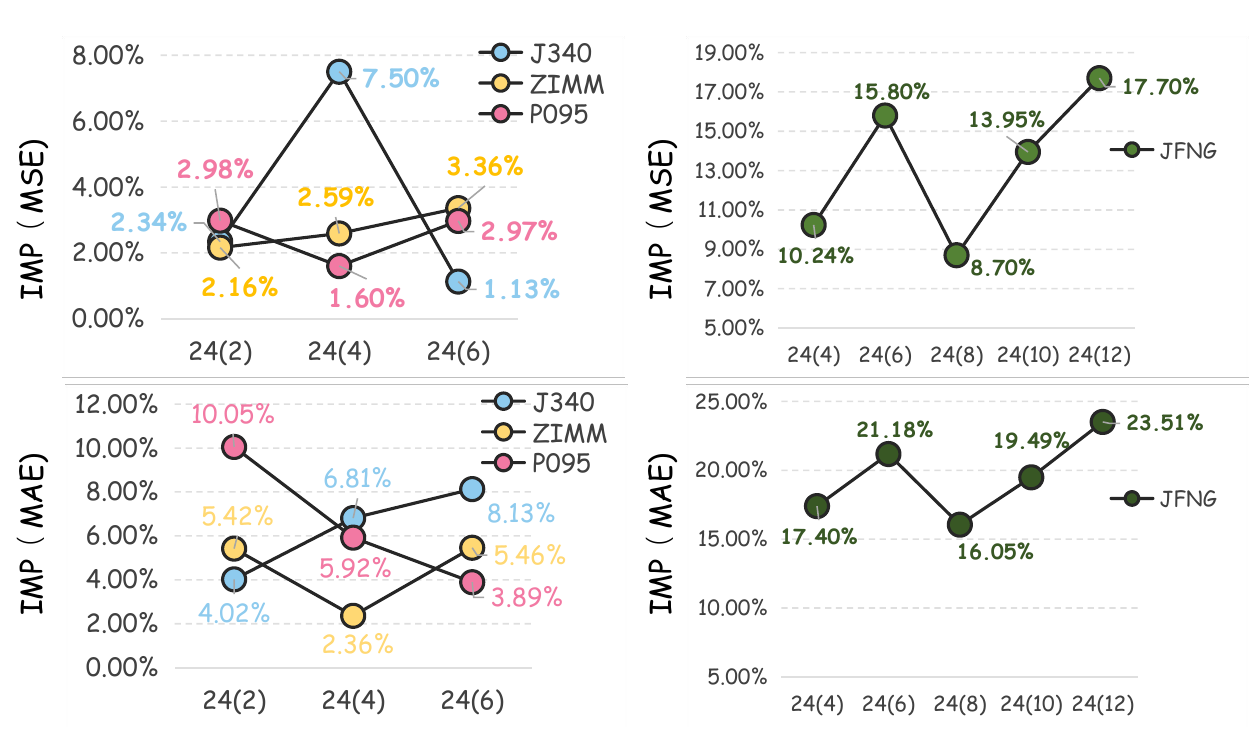}
\caption{The variation trends of MSE and MAE improvement rates (IMP\%) across different stations as the prediction length increases. The horizontal axis represents the forecasting horizon, while the vertical axis shows the relative improvement percentage.} 
\label{fig_multi-scale}
\end{figure}

A key observation is that the relative performance gain of RainBalance tends to persist or even increase with longer forecast horizons.
This trend is especially evident at the JFNG site, where the average MSE improvement (IMP\%) increased steadily from 17.40\% to 23.51\% as the prediction step grew from 4 to 12.
This indicates that RainBalance is not merely a short-term correction module, but also crucial for maintaining accuracy in long-term predictions.
It effectively mitigates common issues such as error accumulation and distribution shift that typically degrade long-range forecasting.
For the other stations, the method maintained consistent improvements across all forecast horizons, confirming that the balancing mechanism of RainBalance is temporally robust and does not decay over time.
\subsubsection{Multi-Forecast Resolution Evaluation}
Our experimental datasets naturally encompass two temporal resolutions: the JFNG station provides high-resolution observations at 15-minute intervals, while the J340, ZIMM, and P095 stations contain standard 1-hour resolution data. This setup offers an ideal condition for examining the effectiveness of our method under different temporal granularities.

Experimental results clearly demonstrate that RainBalance exhibits a greater advantage when processing high-resolution precipitation data. As noted earlier, the JFNG station (15-minute interval) achieves the most significant performance improvement. We attribute this to the inherent characteristics of high-resolution data, which contain richer short-term evolution patterns and more intense instantaneous fluctuations. These properties introduce stronger inter-variable interactions, shorter temporal dependencies, and more pronounced “dual imbalance” phenomena in the data. The clustering and probabilistic distribution modeling mechanisms of RainBalance are particularly adept at identifying and leveraging meaningful patterns in such high-dimensional, dynamic, and imbalanced environments, thereby yielding substantial performance gains.

In contrast, data with a 1-hour temporal resolution are relatively smoother and less volatile, making the forecasting task inherently easier. Consequently, the improvement brought by RainBalance remains consistent but less pronounced.

In conclusion, the performance of RainBalance is positively correlated with the temporal resolution of the data. It proves especially effective for high-resolution precipitation nowcasting, highlighting its strong capability to model complex and rapidly evolving meteorological processes.
\subsubsection{Extreme precipitation Evaluation}
To evaluate the efficacy of RainBalance in extreme precipitation forecasting, we conducted specialized experiments using the top-performing xLSTM model on the JFNG station, focusing on extreme rainfall events. 
We selected only the JFNG station for this analysis because its \textbf{15-minute temporal resolution} allows capturing short-duration, intense rainfall events, whereas the 1-hour resolution at other stations may smooth out peak values. 
This ensures a more accurate evaluation of the model’s performance under extreme rainfall conditions. 

We followed the T/CMSA 0013-2019 standard\footnote{\url{http://www.chinamsa.org/uploads/file/20191106142922_61962.pdf}}, 
under which extreme rainfall is defined as any hourly period with precipitation exceeding 8 mm. As shown in Table \ref{tab:Ours_extreme_comparison}, \textit{RainBalance} demonstrates remarkable performance improvements in predicting extreme precipitation across various forecasting horizons.
\begin{table}[h]
\setlength\tabcolsep{2.5pt}
\centering
\caption{Comparison of base models with and without the proposed module on extreme rainfall events. Values in `+Ours` that improve over the base model are shown in \textbf{bold}. IMP(\%) indicates the percentage improvement of +Ours over the base model.}
\resizebox{0.98\linewidth}{!}{
\begin{tabular}{c|c|cccc|cc}
\toprule
\multirow{2}{*}{\textbf{Dataset}} &
\multirow{2}{*}{\textbf{Horizons}} &
\multicolumn{2}{c}{\textbf{xLSTM}} & 
\multicolumn{2}{c|}{\textbf{+Ours}} &
\multicolumn{2}{c}{\textbf{IMP(\%)}} \\ 

 & & MSE$\downarrow$ & MAE$\downarrow$ & MSE$\downarrow$ & MAE$\downarrow$ & MSE$\uparrow$ & MAE$\uparrow$  \\ 
\midrule

\multirow{5}{*}{\rotatebox[origin=c]{90}{\textbf{JFNG}}}
 & 4  & 2.7952 & 1.1286 & \textbf{2.4181} & \textbf{1.0234}  & 13.49 & 9.34 \\
 & 6  &  8.1605& 2.3553& \textbf{7.9965} & 2.4622 & 2.01& -4.54 \\
 & 8  & 16.4365 & 3.4149 &\textbf{11.6083} &\textbf{2.6326} & 29.38 & 22.93 \\
 & 10 &22.4399 &4.0996 & \textbf{14.6040} & \textbf{2.7005} &34.92& 34.13 \\
 & 12 & 24.5222 & 4.3331 & \textbf{23.5603 }&\textbf{3.8398} &3.93 &11.40 \\
\bottomrule
\end{tabular}}
\label{tab:Ours_extreme_comparison}
\end{table}

The proposed module achieves particularly substantial gains in medium to long-term forecasts (horizons 8-12), with MSE reductions of 29.38\% and 34.92\% at horizons 8 and 10, respectively. Correspondingly, MAE improvements reach 22.93\% and 34.13\% at these horizons, indicating not only reduced large errors but also consistent enhancement in overall prediction accuracy. This trend underscores RainBalance's effectiveness in mitigating error accumulation in extended forecasting scenarios.

Although most horizons exhibit consistent improvements, the slight MAE degradation (-4.54\%) observed at horizon 6 reflects a natural trade-off between different optimization objectives. Our method is designed to suppress large prediction errors in extreme precipitation events, as evidenced by the consistent improvement in MSE across all horizons. Since MSE is more sensitive to extreme values whereas MAE weights all errors equally, this pattern underscores our method’s effectiveness in tackling the key challenge of extreme precipitation forecasting: mitigating high-impact errors in the most critical events.

\section{Conclusion}
In this paper, we identify and formally define the Dual-Imbalance problem inherent in precipitation nowcasting, a challenge characterized by the co-existence of zero-inflation and extreme value scarcity. To address this, we propose RainBalance, a novel, plug-and-play module that reformulates the forecasting task from learning imbalanced single-value labels to modeling continuous cluster probability distributions. By synergistically integrating sample clustering with a variational autoencoder, RainBalance effectively mitigates the learning bias in data-driven models, enabling them to capture both frequent non-rainfall patterns and rare but critical extreme events more effectively.

Extensive experiments on four real-world GNSS stations with diverse geographical and climatic characteristics demonstrate the robustness and generalizability of our approach. RainBalance consistently enhances the performance of multiple state-of-the-art forecasting models across different architectural paradigms, achieving significant improvements. The notable performance gains in extreme precipitation forecasting further underscore its practical value for disaster mitigation and early warning systems. We believe RainBalance provides a principled and effective solution to a fundamental challenge in meteorological AI, opening up new avenues for reliable precipitation nowcasting.
\section*{Acknowledgment}
This work was in part supported by the Science and Technology Innovation 2030 (Grant No.2022ZD0160604), NSFC (Grant No.62176194) and Key R\&D Program of Hubei Province (Grant No.2023BAB083).

\bibliography{reference}
\bibliographystyle{IEEEtran}

\end{document}